\documentclass{article}

\usepackage{microtype}
\usepackage{graphicx}
\usepackage{subfigure}
\usepackage{booktabs} 
\usepackage{fancybox}

\usepackage{import}

\usepackage{hyperref}

\usepackage[accepted]{icml2019}
\usepackage[nonatbib]{} 

\usepackage{tikz}
\usetikzlibrary{shapes, arrows, positioning, decorations.markings}

\definecolor {processblue}{cmyk}{0.96,0,0,0}
\tikzstyle{int}=[draw, fill=blue!20, minimum size=2em]
\tikzstyle{init} = [pin edge={to-,thin,black}]

\usetikzlibrary{shapes}
\usetikzlibrary{fit}
\usetikzlibrary{chains}
\usetikzlibrary{arrows}

\tikzset{
semi/.style={
  semicircle, ,top color =white , bottom color = processblue!20 ,
draw, processblue , text=blue,
  draw,
  minimum size=0.3cm
  }
}

\tikzstyle{plate} = [draw, rectangle, rounded corners, fit=#1]
\tikzstyle{wrap} = [inner sep=0pt, fit=#1]

\tikzstyle{caption} = [node distance=0] %
\tikzstyle{bottom plate caption} = [caption, node distance=0, inner sep=0pt,
below left=-5pt and 0pt of #1.south east] %

\tikzstyle{top plate caption} = [caption, node distance=0, inner sep=0pt,
below left=0pt and 0pt of #1.north east] %

\newtheorem{thm}{Theorem}

\usepackage{times}
\usepackage{url}
\usepackage{latexsym}
\usepackage{graphicx}
\usepackage{epsfig}
\usepackage{algorithm}
\usepackage{algorithmic}
\usepackage{amsmath}
\usepackage{caption}
\usepackage{epstopdf}
\usepackage{amsfonts}
\usepackage{amssymb}
\usepackage{color}

\usepackage{pdfpages}

\newcommand{\diba}{\overline{d}_{\mathrm{IB}}}

\newcommand{\rd}{R_{\rm IB-RD}}

\icmltitlerunning{Aggregated Learning:  A Deep Learning Framework Based on Information-Bottleneck Vector Quantization}

\begin{document}
\twocolumn[
\icmltitle{Aggregated Learning:  A Deep Learning Framework Based on Information-Bottleneck Vector Quantization}




\begin{icmlauthorlist}
\icmlauthor{Hongyu Guo}{nrc}
\icmlauthor{Yongyi Mao}{ou}
\icmlauthor{Ali Al-Bashabsheh}{bh}
\icmlauthor{Richong Zhang}{bh}
\end{icmlauthorlist}

\icmlaffiliation{nrc}{National Research Council Canada, 
  \texttt{hongyu.guo@nrc-cnrc.gc.ca}}
\icmlaffiliation{ou}{
University of Ottawa, \texttt{yymao@eecs.uottawa.ca}}
\icmlaffiliation{bh}{
   Beihang University, Beijing, China, \texttt{entropyali@gmail.com, zhangrc@act.buaa.edu.cn}}


\icmlkeywords{Machine Learning, ICML}

\vskip 0.3in
]



\printAffiliationsAndNotice{} 

\begin{abstract}
    Based on the notion of information bottleneck (IB), we formulate a quantization problem called ``IB quantization". We show that IB quantization is equivalent to learning based on the IB principle. Under this equivalence, 
	the standard neural network models can be viewed as scalar (single sample) IB quantizers.  It is known, from conventional rate-distortion theory, that scalar quantizers are inferior to vector (multi-sample) quantizers.  Such a deficiency then inspires us to develop a novel learning framework, AgrLearn, that corresponds to vector IB quantizers for learning with neural networks.  Unlike standard networks, AgrLearn simultaneously optimizes against multiple data samples. We experimentally verify that AgrLearn can result in significant improvements when applied to several current deep learning architectures for image recognition and text classification. We also empirically show that AgrLearn can  reduce up to 80\% of the training samples needed for ResNet training.
\end{abstract}

\section{Introduction}
The revival of neural networks in the paradigm of deep learning \cite{LeCunBH15:DeepLearning} has stimulated intense interest in understanding the working of deep neural networks (e.g. \cite{Tishby17:IB-deepLearning,ZhangBHRV16:rethinkingGeneralization}). Among various research efforts, an information-theoretic approach, {\em information bottleneck} (IB) \cite{Tishby99:IB}, stands out as a promising and fundamental tool to theorize the learning of deep neural networks \cite{Tishby17:IB-deepLearning,michael2018on,DBLP:conf/icml/DaiZGW18,DBLP:conf/icml/BelghaziBROBHC18}. 

This paper builds upon some previous works on IB~\cite{Tishby03:IB-rate-distortion,Tishby10:IB-LearningGeneralization,Tishby99:IB}.  
Under the IB principle, the objective of learning is to find a representation, or bottleneck, $T$ of an example $X$ so that the mutual information between $X$ and $T$ is minimized whereas the mutual information between $T$ and the class label $Y$ is maximized. This results in a constrained optimization problem, which we refer to as the {\em IB learning} problem. In this paper, we introduce an unconventional quantization problem, which we refer to as {\em IB quantization}. We prove that the objective of IB
quantization, namely, designing quantizers that achieve the rate-distortion function, is equivalent to the objective of IB learning, thereby establishing an equivalence between the two problems.

Under this equivalence, one can regard the current neural network models as ``scalar IB quantizers''. In the literature of quantization and rate-distortion theory~\cite{Shannon59:rateDistortion}, it is well known that scalar quantizers are in general inferior to vector quantizers. This motivates us to adopt a vector quantization approach to learning with deep neural networks.  The main proposal of this paper is a simple framework for neural network modeling, which  we call {\em Aggregated Learning} (AgrLearn).  

Briefly, in AgrLearn, a neural network classification model is structured to simultaneously classify $n$ objects (Figure \ref{fig:framework}). This resembles a standard vector quantizer, which simultaneously quantizes multiple signals. In the training of the AgrLearn model, $n$ random training objects are aggregated to a single amalgamated object and passed to the model.  When using the trained model for prediction, the input to the model is also an aggregation of $n$ objects, which can be all different or replicas of the object.  

\begin{figure}
\centering
\includegraphics[width=\columnwidth]{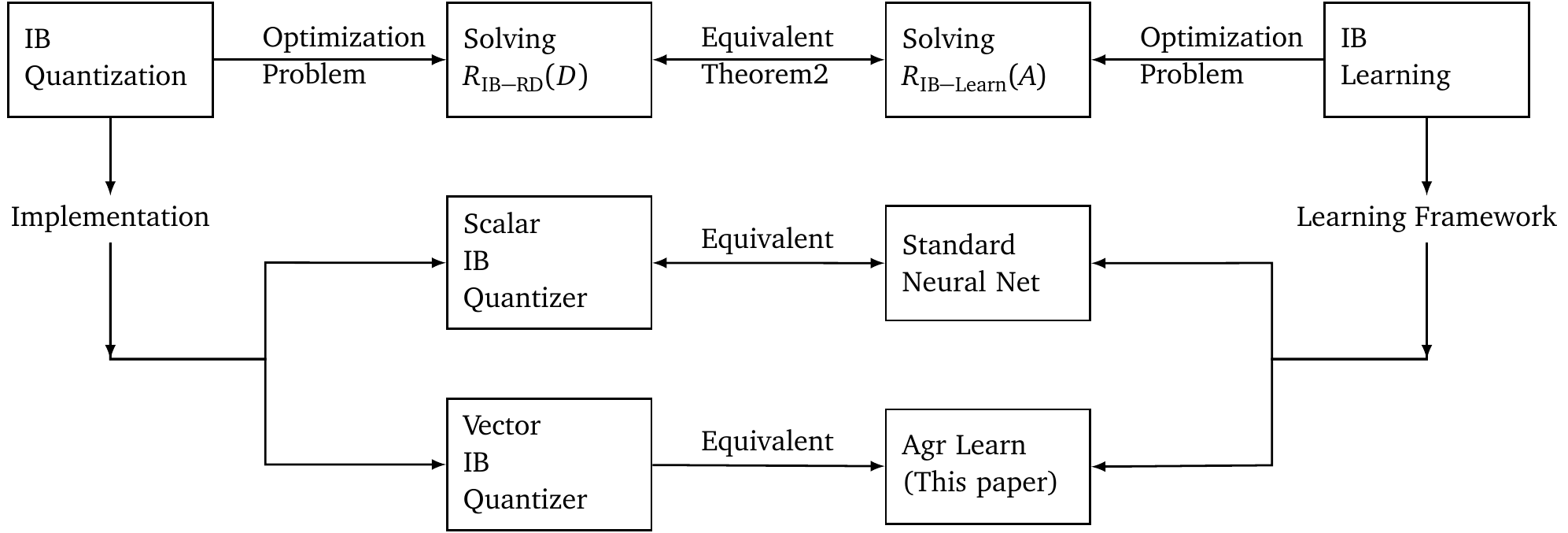}
\caption{The relationship between IB quantization, IB learning, standard neural networks and the proposed AgrLearn framework.}
\label{fig:equivalence}
\end{figure}

We conduct extensive experiments 
applying AgrLearn to the current art of deep learning architectures for image and text classification. Experimental results suggest that AgrLearn brings significant gain in classification accuracy.  

In practice, AgrLearn can be easily integrated into existing neural network architectures with just addition of a few lines of code. Furthermore our experiments suggest that AgrLearn can dramatically reduce the required training examples (e.g., by 80\% on Cifar10). 
This can be particularly advantageous for real-world learning problem with scarce labeled data. 

Our main contributions can be summarized as follows.
\begin{itemize}
\item  We formulate the IB quantization problem and prove that
it is equivalent to learning under the IB principle. 
\item Under this equivalence, standard neural network models can be regarded as scalar IB quantizers. Recognizing theoretical inferiority of scalar quantizers to vector quantizers, we devise a novel neural-network learning framework, AgrLearn, that is equivalent to vector IB quantizers. 
\item We empirically demonstrate that AgrLearn, when plug into a model with virtually no programming or tuning effort, can significantly improve the classification accuracy of recent deep models in both image recognition and text classification.
\item We experimentally show that AgrLearn can  dramatically reduce the amount of  training data needed for network learning. 
\end{itemize}

\section{The Aggregated Learning Framework}

\subsection{Information Bottleneck Learning}

Throughout the paper, a random variable will be denoted by a capitalized letter, e.g., $X$, and a value it may take is denoted by its lower-case version, e.g., $x$. The distribution of a random variable is denoted by $p$ with a subscript indicating the random variable, e.g., $p_X$. 

We consider a generic classification setting. We use ${\cal X}$ to denote the space of {\em objects} to be classified, where an object $X$ is distributed according to an {\em unknown} distribution $p_X$ on ${\cal X}$. Let ${\cal Y}$ denote the space of class labels. There is an unknown function $F:{\cal X}\rightarrow {\cal Y}$ which assigns an object $X$ a class label $Y:=F(X)$.  Let ${\cal D}_{\cal X}:=\{x_1, x_2, \ldots, x_N\}$ be a given set of {\em training examples},  drawn i.i.d. from $p_X$. The objective of learning in this setting is to find an approximation of $F$ based on the {\em training set} ${\cal D}:=\{(x_i, F(x_i): x_i\in {\cal D}_{\cal X}\}$.

In the {\em information bottleneck} (IB) formulation \cite{Tishby99:IB} of such a learning problem, one is interested in learning a representation $T$ of $X$ in another space ${\cal T}$. We will refer to $T$ as a {\em bottleneck} representation. When using a neural network model for classification, in this paper, we regard $T$ as the vector computed at the {\em final hidden layer} of the network before it is passed to a standard soft-max layer to generate the predictive distribution over ${\cal Y}$. We will denote by $h$ the function implemented by the network that computes $T$ from the $X$, namely, $h$ corresponds to the part of the network from input all the way to the final hidden layer and $T=h(X)$. 

The IB method insists on the following two principles.

I. The mutual information $I(X; T)$ should be as small as possible. \\
II. The mutual information $I(Y; T)$ should be as large as possible.

Principle I insists that $h$ squeezes out the maximum amount of information contained in $X$ so that the information irrelevant to $Y$ is removed when constructing $T$.  Principle II insists that $T$ contains the maximum amount of information about $Y$ so that all the information relevant to classification is maintained. Intuitively, the first principle forces the model not to over-fit due to the irrelevant features of $X$, whereas the second aims at maximizing the classification accuracy. 

In general the two principles have conflicting objectives. A natural approach to deal with this conflict is to set up a constrained optimization problem which implements one principle as the objective function and the other as the constraint. This results in the following {\em IB learning problem} , where $T$ is random variable taking values in ${\cal T}$ and forming a Markov chain $Y-X-T$ with $(Y, X)$. 

\centerline{
\shadowbox
{
\parbox[t]{7.5cm}{
\noindent {\bf The IB Learning Problem}
\begin{multline}
\label{eq:IB-learn-problem}
\widehat{p}_{T|X}:=\arg\min_{p_{T|X}: I(Y; T)\ge A} I(X; T)  \\
{\rm ~for ~some ~prescribed ~value} ~A
\end{multline}
and we define
\begin{align}
R_{\rm IB-learn}(A):=\min_{p_{T|X}: I(Y; T)\ge A} I(X; T)
\label{eq:r-learn}
\end{align}
}
}
}

Here we have assumed that {\em the joint distribution $p_{XY}$ is known}. We will adopt this assumption for now and later discuss the realistic setting in which one only has access to the empirical observation of 
$p_{XY}$ through training data.

It is worth noting that the IB learning problem in (\ref{eq:IB-learn-problem}) is in fact more general than learning the bottleneck $T$ in a neural network. This latter problem, which we refer to as the {\em Neural Network IB (NN-IB) learning problem}, restricts that $T$ depends on $X$ {\em deterministically}:

\centerline{
\shadowbox
{
\parbox[t]{7.5cm}{
\noindent {\bf The NN-IB Learning Problem}
\begin{equation}
\label{eq:IB-NN}
\widehat{h}:=\arg\min_{h: I(Y; T)\ge A} I(X; T)
\end{equation}
where $h$ ranges over the set of all functions mapping ${\cal X}$ to ${\cal T}$, and we define
\[
R_{\rm IB-NN}(A):= \min_{h: I(Y; T)\ge A} I(X; T)
\]
}
}
}

For any given $p_{XY}$, bottleneck space ${\cal T}$ and non-negative value $A$, it is apparent that 
\begin{equation}
\label{eq:IBNN_isSuboptimal}
R_{\rm IB-learn}(A) \le R_{\rm IB-NN}(A).  
\end{equation}
Furthermore, unless one delicately constructs
$p_{XY}$ and ${\cal T}$, in general, inequality (\ref{eq:IBNN_isSuboptimal}) holds strictly. 
This fact suggests that the conventional neural network, which uses a deterministic mapping to transform the input to latent representation is in general sub-optimal for solving the IB learning problem (\ref{eq:IB-learn-problem}). 

In a recent work \cite{DVIB:AlemiFD016}, stochastic mappers are introduced to neural networks and the networks are trained by minimizing a loss defined using information bottleneck, an  optimization problem equivalent to (\ref{eq:IB-learn-problem}). 
This in effect turns problem (\ref{eq:IB-NN}) to problem (\ref{eq:IB-learn-problem}). Since the new optimization is intractable, the authors of \cite{DVIB:AlemiFD016} propose a variational approximation to the problem and show, even under such an approximation, that stochastically mapping the input $X$ to a bottleneck $T$ results in a performance superior to using the 
deterministic mappings as is in standard neural networks. This observation is consistent with the sub-optimality of the NN-IB learning formulation (\ref{eq:IB-NN}) suggested in inequality (\ref{eq:IBNN_isSuboptimal}), although the thrust of \cite{DVIB:AlemiFD016} is not explicitly concerned with improving upon such sub-optimality. 

This paper deals with the sub-optimality of the NN-IB learning problem relative to the IB-learning problem  (\ref{eq:IBNN_isSuboptimal}) from a different perspective. The perspective is developed by identifying a quantization problem, which we call {\em IB quantization}, associated with the IB learning problem (\ref{eq:IB-learn-problem}).




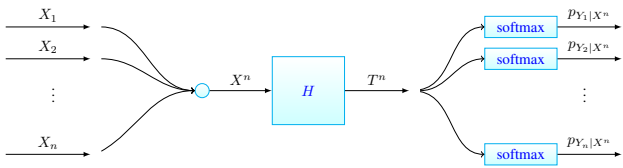
\begin{figure}[ht!]
\centerline{
\scalebox{0.565}{
\begin{tikzpicture}[-latex ,auto ,node distance =2 cm and 2 cm ,on grid ,
semithick ,
state/.style ={ circle ,top color =white , bottom color = processblue!20 ,
draw, processblue , text=blue , minimum width =0.1cm},
box/.style ={rectangle ,top color =white , bottom color = processblue!20 ,
draw, processblue , text=blue , minimum width =1.7cm , minimum height = 0.5cm},
highbox/.style ={rectangle ,top color =white , bottom color = processblue!20 ,
draw, processblue , text=blue , minimum width =1.7cm , minimum height = 1.6cm},
neuron/.style ={rectangle ,top color =white , bottom color = red!20 ,
draw, red , text=red , minimum width =2.6cm , minimum height = 3.6cm, rounded corners},
triangle/.style = {top color =white , bottom color = processblue!20 ,
draw, processblue , text=blue, regular polygon, regular polygon sides=3, minimum size=0.5cm, draw },
node rotated/.style = {rotate=270},
    border rotated/.style = {shape border rotate=270}]

\node[](realcenter){};
\node[](center)[above=0.2cm of realcenter]{};
\node[highbox](cnn)[above=0.5cm of center]{$H$};

\node[state](sold)[left=2.5cm of cnn]{};
\path (sold) edge [] node[]{$X^n$} (cnn);
\node[](lcenter0)[left=2.5cm of sold]{};
\node[](lcenter1)[above=1.5cm of lcenter0]{};
\node[](lcenter2)[above=0.75cm of lcenter0]{};
\node[](lcentern)[below=1.5cm of lcenter0]{};

\node[](lcenter11)[left=2.25cm of lcenter1]{};
\node[](lcenter21)[left=2.25cm of lcenter2]{};
\node[](lcentern1)[left=2.25cm of lcentern]{};

\path (lcenter11) edge [] node[]{$X_1$}  (lcenter1);
\path (lcenter21) edge [] node[]{$X_2$} (lcenter2);
\path (lcentern1) edge [] node[]{$X_n$} (lcentern);
\node[]()[left=1cm of lcenter0]{$\vdots$};

\draw [->] (lcenter1) to [out=0,in=180] (sold);
\draw [->] (lcenter2) to [out=0,in=180] (sold);
\draw [->] (lcentern) to [out=30,in=180] (sold);

\node[](snew)[right=2.5cm of cnn]{};
\node[](rcenter0)[right=2.5cm of snew]{};
\node[box](rcenter1)[above=1.5cm of rcenter0]{softmax};
\node[box](rcenter2)[above=0.75cm of rcenter0]{softmax};
\node[box](rcentern)[below=1.5cm of rcenter0]{softmax};

\node[](rcenter11)[right=2.5cm of rcenter1]{};
\node[](rcenter21)[right=2.5cm of rcenter2]{};
\node[](rcentern1)[right=2.5cm of rcentern]{};
\node[]()[right=1.5cm of rcenter0]{$\vdots$};

\path (cnn) edge [] node[]{$T^n$}  (snew);

\draw [->] (snew) to [out=10,in=180] (rcenter1);
\draw [->] (snew) to [out=0,in=180] (rcenter2);
\draw [->] (snew) to [out=-10,in=180] (rcentern);

\path (rcenter1) edge [] node[]{$p_{Y_1|X^n}$} (rcenter11);
\path (rcenter2) edge [] node[]{$p_{Y_2|X^n}$} (rcenter21);
\path (rcentern) edge [] node[]{$p_{Y_n|X^n}$}(rcentern1);











\end{tikzpicture}
}
}
\caption{The Aggregated Learning (AgrLearn) framework. The small circle denotes concatenation.}
\label{fig:framework}
\vspace{-0.5cm}
\end{figure}

\subsection{The IB Quantization Problem}

We now formulate the {\em IB quantization problem}. We note that this problem was first identified in \cite{Tishby03:IB-rate-distortion}.

To begin, let 
$(X_1, Y_1), (X_2, Y_2), \ldots, (X_n, Y_n)$ be drawn i.i.d. from $p_{XY}$, in which $X_1, X_2, \ldots, X_n$ serve as an {\em information source}. The sequence 
$(X_1, X_2, \ldots, X_n)$ is also denoted by $X^n$, and likewise $(Y_1, Y_2, \ldots, Y_n)$ by $Y^n$. 


Let the bottleneck space ${\cal T}$ be given.  
An $(n, 2^{nR})$ {\em IB-quantization code} is a pair $(f_n, g_n)$ of functions, in which $f_n:{\cal X}^n\rightarrow \{1, 2, \ldots, 2^{nR}\}$ maps each sequence in ${\cal X}^n$ to an integer in $\{1, 2, \ldots, 2^{nR}\}$ and $g_n:\{1, 2, \ldots, 2^{nR}\} \rightarrow {\cal T}^n$ maps an integer in $\{1, 2, \ldots, 2^{nR}\}$ to a sequence in ${\cal T}^n$. Using this code, $f_n$ encodes the sequence $X^n$ as the integer $f_n(X^n)$, and $g_n$ ``reconstructs'' $X^n$ as a representation $T^n:=(T_1, T_2, \ldots, T_n):= g_n(f_n(X^n))$ in ${\cal T}^n$.   Using the standard nomenclature in quantization, the quantity $R$ is referred to as the {\em rate} of the code, and $n$ as the 
{\em length} of the code.

Define the distortion between $x\in {\cal X}$ and $t\in {\cal T}$, with respect to any conditional distributions $q_{Y|X}$ and $q_{Y|T}$, as 
\begin{equation}
d_{\rm IB}(x, t;q_{Y|X}, q_{Y|T}):={\rm KL}(q_{Y|X}(\cdot|x) || q_{Y|T}(\cdot|t)),
\end{equation}
where ${\rm KL}(\cdot || \cdot)$ denotes the KL divergence.

Note that the code $(f_n, g_n)$ induces a joint distribution over the
Markov chain $Y^n-X^n-T^n$. Under this joint distribution, the conditional distributions  $p_{Y_i|X_i}$ and  $p_{Y_i|T_i}$ are well defined for each $i=1, 2, \ldots, n$. Then for every two sequences $x^n\in {\cal X}^n$ and $t^n\in {\cal T}^n$,  we define their {\em IB distortion} as
\begin{align}
	\overline{d}_{\rm IB}(x^n, t^n): = \frac{1}{n}\sum\limits_{i=1}^n d_{\rm IB}(x_i, t_i; p_{Y_i|X_i}, p_{Y_i|T_i}).
	\label{eq:dist-seq}
\end{align}

Under these definitions, the {\em IB quantization problem} is to find a code $(f_n,
g_n)$ having the smallest rate $R$ subject to the constraint $E \overline{d}_{\rm IB}(X^n, T^n) \le D$,
where $E$ denotes expectation.
More precisely,
given $p_{XY}$ and ${\cal T}$,
a rate distortion pair $(R,D)$ is said to be \emph{achievable } if
\begin{align}
	\label{eq:achievable}
	E\diba(X^{n},T^{n}) \leq D
\end{align}
for some sequence $(f_n,g_n)$ of $(2^{nR},n)$ codes and $n\rightarrow \infty$. The \emph{rate-distortion function} $\rd(D)$ is the smallest rate
$R$ s.t. $(R,D)$ is achievable.
Note that different from the distortion function in a conventional
quantization problem, see, e.g., \cite{Cover06:InfoTheory}, the IB distortion $\overline{d}_{\rm IB}$
in fact depends on the choice of the IB-quantization code $(f_n, g_n)$~\cite{Tishby03:IB-rate-distortion}.

\begin{thm} 
\label{thm:main}
{\em Given $p_{XY}$ and ${\cal T}$, the IB rate-distortion function can be written as 
\begin{equation}
\label{eq:IB-RD-problem}
R_{\rm IB-RD}(D) = \min_{p_{T|X}: E{d}_{\rm IB}(X; T)\le D} I(X; T),
\end{equation}
where the expectation is over the Markov chain $Y-X-T$ specified by $p_{XY}$ and $p_{T|X}$.
}
\end{thm}

This theorem provides a limit on the quantization rate, in terms of the mutual information, below which no code exists without violating the distortion requirement.
The theorem was first shown in \cite{Tishby03:IB-rate-distortion}, where it was
assumed that $|\mathcal{T}| \geq |\mathcal{X}|+2$. We remark that this assumption is not necessary and 
provide a proof in the supplementary material.

In the literature of quantization, a quantization code of length $1$ is referred to as a {\em scalar quantizer} whereas a quantization code having length greater than $1$
is called a {\em vector quantizer}.  In general, to achieve the rate-distortion limit $R_{\rm IB-RD}(D)$ in Theorem \ref{thm:main}, scalar quantizers are insufficient. One must rely on vector quantizers to approach (or achieve) this limit.



\subsection{IB Learning as IB Quantization}
The form of the rate-distortion function $R_{\rm IB-RD}$ of IB quantization given in Theorem \ref{thm:main} resembles greatly the optimal objective $R_{\rm IB-learn}$
achieved in IB learning. In fact, simple manipulation of the constraints of
(\ref{eq:IB-learn-problem}) and (\ref{eq:IB-RD-problem}) gives rise to the following theorem.
\begin{thm} 
$R_{\rm IB-learn}(A)=R_{\rm IB-RD}(I(X; Y)-A).$
\end{thm}
This theorem suggests that solving the IB learning problem that achieves 
$R_{\rm IB-learn}(A)$ is equivalent to finding the optimal IB-quantization code that achieves 
$R_{\rm IB-RD}(I(X; Y)-A)$ (noting that given $p_{XY}$, $I(X; Y)$ is merely a given constant). When viewing IB learning as IB quantization, it is evident that the standard neural network that uses a deterministic mapping $h:{\cal X}\rightarrow {\cal T}$ is equivalent to a scalar IB quantizer $(f_1, g_1)$, where $h$ is the composition  $g_1\circ f_1$ of $g_1$ and $f_1$. The sub-optimality of 
scalar quantizers for IB quantization  (or equivalently the sub-optimality of the NN-IB learning approach to IB learning) then motivates us to move away from the standard neural network framework and consider using vector IB quantizers. This gives rise to the {\em Aggregated Learning} framework, which is presented next.  Before proceed, we wish to stress that the proposal of AgrLearn is strongly justified by the above theoretical development. To recapitulate, the relationship between IB learning, IB quantization, standard neural networks and the AgrLearn framework is summarized in Figure  \ref{fig:equivalence}.


\subsection{Aggregated Learning (AgrLearn)}
\label{evalPro}
We now present the framework of Aggregated Learning, or AgrLearn in short, for neural networks. 

Instead of taking a single object in ${\cal X}$ as input, AgrLearn takes $n$ objects $X^n$ as input. Here the value $n$ is prescribed by the model designer and is referred to as the {\em fold} of AgrLearn.  In fold-$n$ AgrLearn  (Figure \ref{fig:framework}), the bottleneck is generated as $T^n=H(X^n)$ using some function $H$, and this ``aggregated bottleneck'' $T^n$ is then passed to $n$ parallel soft-max layers. 
Let ${\bf softmax}(y_i|H(x^n); \theta_i)$ denote the $i^{\rm th}$ softmax layer, with learnable parameter $\theta_i$, which  maps the bottleneck $H(x^n)$ to the predictive distribution $p_{Y_i|X^n}$ of label $Y_i$. The AgrLearn model then states
\begin{multline}
\label{eq:aggreLearn}
p_{Y^n|X^n}(y^n|x^n)=\prod\limits_{i=1}^n p_{Y_i|X^n}(y_i|x^n) \\
=\prod\limits_{i=1}^n {\bf softmax}(y_i|H(x^n); \theta_i).
\end{multline}


To train the AgrLearn model, a number of aggregated training examples are formed (as many as one can), each by concatenating $n$ random objects sampled from ${\cal D}_{\cal X}$ with replacement. We then minimize the cross-entropy loss over all aggregated training examples. 

We note that in an ideal world, instead of minimizing the cross-entropy loss, one should truthfully optimize a (Lagrangian) formulation of the mutual information objective in (\ref{eq:IB-learn-problem}). This however requires complex computation of mutual information or its approximations, making training prohibitively inefficient. Thus deriving an accurate and efficient approximation of the objective (\ref{eq:IB-learn-problem}) for AgrLearn is of great research interest. 
Nonetheless, in~\cite{Tishby17:IB-deepLearning} Shwartz-Ziv and Tishby show that 
 SGD over cross-entropy loss exhibits well-behaving trajectories in the {\em (I(X; T); I(Y; T))} plane, suggesting that the
 standard cross-entropy loss and the mutual information  objectives are highly correlated, at least in the context of SGD-based training.
Consequently, to this end 
we simply use the cross-entropy loss as a proxy for the true objective in this paper. Such an approach (namely, one using a simple metric to replace the mutual information objective) is in fact often adopted in the practical design of quantizers in the field of data communications (for example, the Lloyd-Max algorithm~\cite{Lloyd:2006:LSQ:2263356.2269955}).

When using the trained model for prediction, the following ``Replicated Classification'' protocol is used\footnote{Two additional protocols are in fact also investigated.
\underline{Contextual Classification}: For each object $X$, $n-1$ random examples are drawn from ${\cal D}_{\cal X}$ and concatenated with $X$ to form the input; the predictive distribution for $X$ generated by the model is then retrieved. Repeat this process $k$ times, and take the average of the $k$ predictive distribution as the label predictive distribution for $X$. \underline{Batched Classification}: Let ${\cal D}_{\cal X}^{\rm test}$ denote the set of all objects to be classified.  In Batched Classification, ${\cal D}_{\cal X}^{\rm test}$ are classified jointly through drawing $k$ random batches of $n$ objects from ${\cal D}_{\cal X}^{\rm test}$. The objects in the $i^{\rm th}$ batch $B_i$ are concatenated to form an input and passed to the model. The final label predictive distribution for each object $X$ in ${\cal D}_{\cal X}^{\rm test}$ is taken as the average of the predictive distributions of $X$ output by the model for all batches $B_i$'s containing $X$.  Since we observe that all three protocols result in comparable performances, all results reported in the paper are obtained using the Replicated Classification protocol.
}. Each object $X$ is replicated $n$ times and concatenated to form the input. The average of $n$ predictive distributions generated by the model is taken as the label predictive distribution for $X$.

\subsection{Complication of Finite Sample Size $N$}
\label{theory}
The analysis above that motivates AgrLearn is based on the assumption that $p_{XY}$ is known. This assumption corresponds to the asymptotic limit of infinite number $N$ of training examples. In such a limit and assuming sufficient capacity of AgrLearn, larger aggregation fold $n$ in theory gives rise to better bottleneck $T$ (in the sense of minimizing $I(X; T)$ subject to the $I(Y; T)$ constraint). 

In practice, one however only has access to the empirical distribution $\widetilde{p}_{XY}$ through observing the $N$ training examples in ${\cal D}_{\cal X}$.  As such, AgrLearn, in the large-$n$ limit, can only solve for the empirical version of the optimization problem 
(\ref{eq:IB-learn-problem}), namely, finding $\min_{p_{T|X}: \widetilde{I}(Y; T)\ge A} \widetilde{I}(X; T)$, where $\widetilde{I}(X; T)$ and $\widetilde{I}(Y; T)$ are $I(X; T)$ and $I(Y; T)$ induced by $\widetilde{p}_{XY}$ and $p_{T|X}$. The solution to the empirical version of the problem in general deviates from that to  the original problem. 
Thus we expect, for finite $N$, that there is a critical value $n^*$ of fold $n$, 
above which AgrLearn degrades its performance with increasing $n$. How to characterize $n^*$ remains open at this time. Nonetheless it is sensible to expect that $n^*$ increases with $N$, since larger $N$ makes $\widetilde{p}_{XY}$ better approximate $p_{XY}$. 

With finite $N$, the product $\left(\widetilde{p}_{XY}\right)^{\otimes n}$ of the empirical distribution $\widetilde{p}_{XY}$ is a non-smooth approximation of the true product $\left({p}_{XY}\right)^{\otimes n}$, and the ``non-smoothness'' increases with $n$. Intuitively, instead of using $\left(\widetilde{p}_{XY}\right)^{\otimes n}$ to approximate $\left({p}_{XY}\right)^{\otimes n}$, using some smoother approximations may improve the performance of 
AgrLearn. In some of our experiments (those in Section \ref{smoother}), we incorporate the strategy of ``MixUp''\cite{mixup17} as a heuristics to smooth $\left(\widetilde{p}_{XY}\right)^{\otimes n}$ in AgrLearn. 

\section{Experimental Studies}
We evaluate AgrLearn with several widely deployed deep network architectures for classification tasks in both image and natural language domains.  Standard benchmarking datasets are used. Unless otherwise specified, fold number $n=8$ is used in all AgrLearn models.

All models examined are trained using mini-batched backprop    for 400 epochs\footnote{Here an epoch refers to going over $N$ {\em aggregated} training examples, where $N=|{\cal D}_{\cal X}|$.} with {\em exactly the same} hyper-parameter settings without dropout. Specifically, weight decay is $10^{-4}$, and each mini-batch contains 64 aggregated training examples. The learning rate is set to 0.1 initially and decays by a factor of 10 after 100, 150, and 250 epochs for all models. 
Each reported performance value (accuracy or error rate) is the median of the performance values obtained in the final 10 epochs. 


\begin{table*}[h]
  \centering
\begin{tabular}{l|c|c|c}\hline
Dataset & ResNet-18& AgrLearn-ResNet-18& Relative Improvement over ResNet-18
\\\hline 
Cifar10& 5.53&4.73 &14.4\%\\ 
Cifar100&25.6 &23.7 & 7.4\%\\ 
SVHN&3.67&3.01&10.6\%\\
\hline \hline
 & WideResNet-22-10 & AgrLearn-WideResNet-22-10& Relative Impr. over WideResNet-22-10 \\\hline 
Cifar10& 3.88&3.24&16.4\%\\ 
Cifar100&19.86&18.18&8.4\%\\ 
SVHN&3.31&2.87&13.0\%\\
\hline
\end{tabular}
  \caption{Test error rates (\%) of ResNet-18, WideResNet-22-10 and their AgrLearn counterparts on Cifar10, Cifar100, and SVHN}
  \label{tab:accuracy:resnet}
\end{table*}

\begin{table*}[h]
  \centering
\begin{tabular}{l|c|c|c}\hline
Dataset & ResNet-18& AgrLearn-ResNet-18& Relative Improvement over ResNet-18
\\\hline 
ImageNet-top1&33.64 &31.61 &5.9\%

\\ 
ImageNet-top5&12.99 &11.51 &11.3\%\\ 
\hline
\end{tabular}
  \caption{Test error rate (\%) comparison of AgrLearn-ResNet-18 and ResNet-18 on ImageNet-2012}
  \label{tab:accuracy:imagenet}
\end{table*}

\subsection{Image Recognition}
Experiments are conducted on the \textbf{Cifar10},   \textbf{Cifar100}, \textbf{SVHN}, and \textbf{ImageNet} datasets with two widely used deep networks architectures, namely ResNet~\cite{DBLP:conf/eccv/HeZRS16} and WideResNet~\cite{DBLP:journals/corr/ZagoruykoK16}.
\vspace{-0.2cm}
\begin{itemize}
\item
\textbf{Cifar10}: this data set has 50,000 training images, 10,000 test images, and  10 image classes. 
\item
\textbf{Cifar100}:  similar to Cifar10 but with  100 classes. 
\item \textbf{SVHN}: the Google street view house numbers recognition data set with 73,257 digits 
(0-9) 32x32 color images for training, 26,032 for testing, and 531,131 additional, easier samples. We did not use the additional images. 
\item
\textbf{ImageNet-2012}: a large image classification dataset~\cite{RussakovskyDSKSMHKKBBF14} with  1.3 million training images, 50,000 validation images, and 1,000 classes.
\end{itemize}
\vspace{-0.2cm}
We apply AgrLearn to  the 18-layer Pre-activation ResNet (``ResNet-18'')~\cite{DBLP:conf/eccv/HeZRS16} implemented in~\cite{liu17}, and the 22-layer Wide\_ResNet (``WideResNet-22-10'')~\cite{DBLP:journals/corr/ZagoruykoK16} as implemented in~\cite{Zagoruyko/code}.

The resulting AgrLearn model
(``AgrLearn-ResNet-18'' and ``AgrLearn-WideResNet-22-10'') differs from ResNet-18 and WideResNet-22-10 only in its $n$ parallel soft-max layers (as opposed to the single soft-max layer in ResNet-18 and WideResNet-22-10). 

\subsubsection{Predictive Performance}
The prediction error rates of AgrLearn-ResNet-18,  AgrLearn-WideResNet-22-10, and ResNet-18 are shown  in Table~\ref{tab:accuracy:resnet}.

It can be seen that AgrLearn significantly boosts the performance of ResNet-18 and WideResNet-22-10. For example, with respect to  ResNet-18, the  relative error reductions are  14.4\%,  7.4\%, and 10.6\% on Cifar10, Cifar100, and SVHN, respectively. On ImageNet, the relative error reductions achieved by AgrLearn-ResNet-18 
for the top1 an top5 error are 5.9\% and 11.3\%, respectively. Similarly significant improvement upon WideResNet is also observed. For example,  
with respect to WideResNet-22-10, the relative error reductions achieved by AgrLearn are 16.4\%, 8.4\%, and 13.0\% on Cifar10, Cifar100, and SVHN, respectively.  Remarkably, this performance gain simply involves plugging  an existing neural network architecture in AgrLearn without any (hyper-)parameter tuning.

\subsubsection{Model Behavior During Training}
\begin{figure}[h]
    \centering
    \includegraphics[height=1.57985234563in]
 {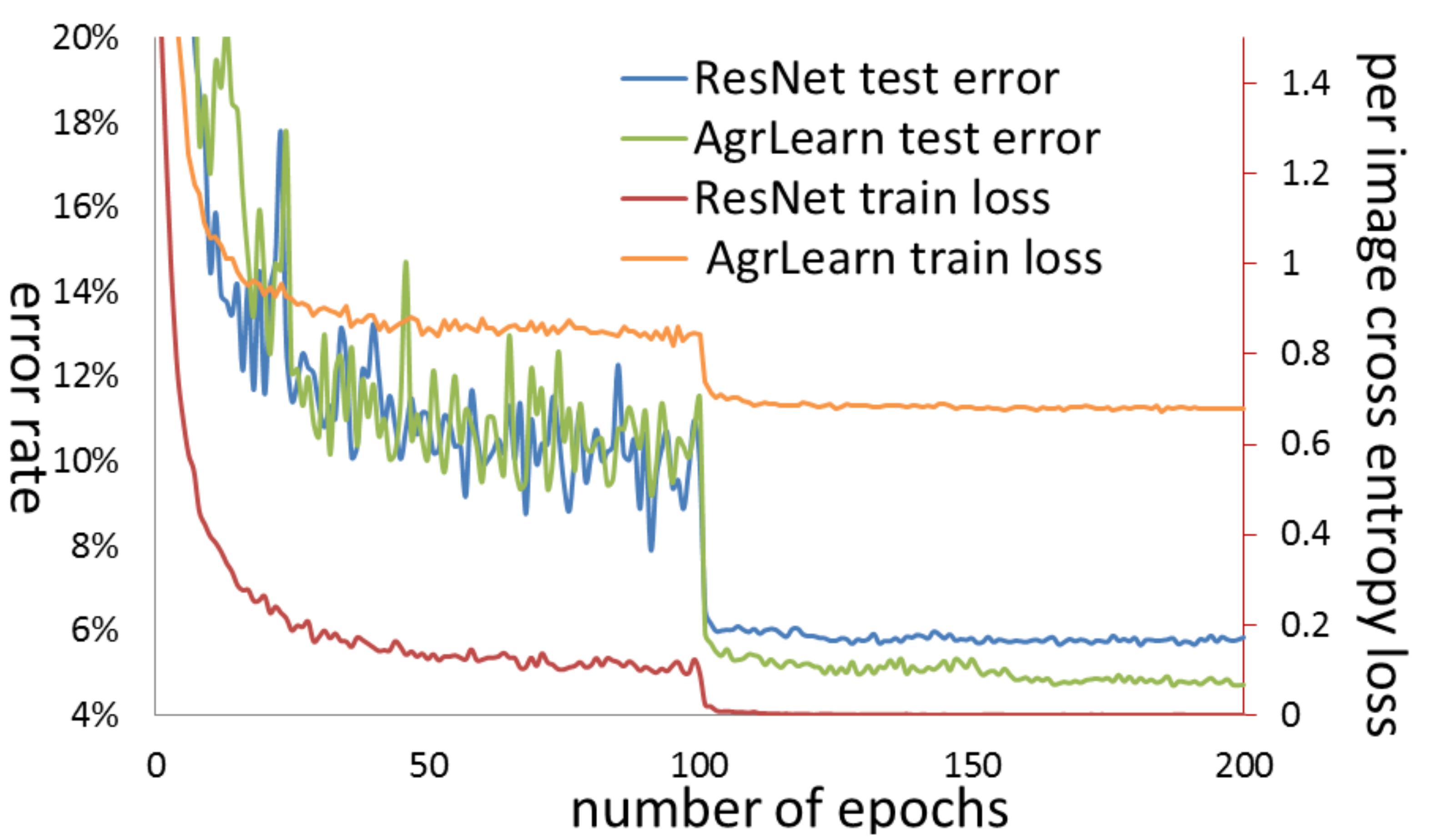}
        \caption{Training loss and test error  on Cifar10.}
        \label{fig:conv}
    \end{figure}

The typical behavior of AgrLearn-ResNet-18 and ResNet-18 (in terms of training cross-entropy loss and testing error rate) across training epochs is shown 
in Figure~\ref{fig:conv}. It is seen that during earlier training epochs,
the test error of Agrlearn (green curve) fluctuates more than that of ResNet (blue curve)
until both curves drop and stabilize. In the ``stable phase'' of training, the test error of AgrLearn continues to decrease whereas the test performance of ResNet 
fails to further improve. This can be explained by the training loss curve of ResNet (red curve),
which drops to zero quickly in this phase and provides no training signal for further tuning 
 the network parameters. In contrast, the training curve of AgrLearn (brown curve) maintains a relatively high level, allowing the model to keep tuning itself. The relatively higher training loss of AgrLearn is due to the much larger space of the amalgamated examples. Even in the stable phase, one expects that the model is still seeing new combinations of images.  In a sense, we argue that aggregating several examples into a single input can be seen as an implicit form of regularization, preventing the model from over-fitted by limited number of individual examples.

\subsubsection{Feature Map Visualization}
\begin{figure}[h]
        \centering
        \includegraphics[height=1.245312346425in, width=2.973in]{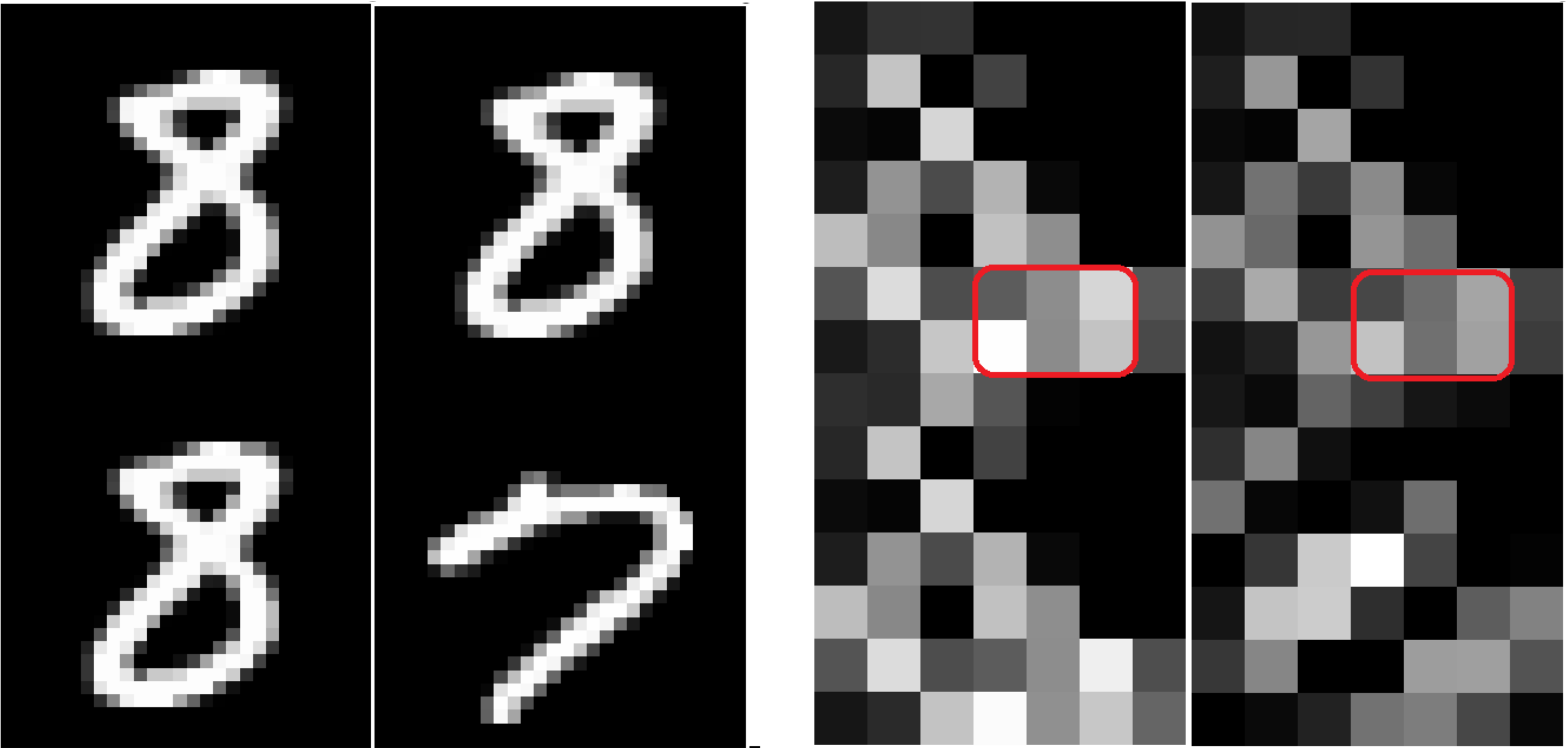}
        \caption{Two inputs and their filter maps on MNIST.}
        \label{vis}
\end{figure}

To visualize the extracted feature by AgrLearn, we perform a small experiment on MNIST using 
a fold-2 AgrLearn on a 3-layer CNN as implemented in~\cite{wu2016tensorpack}. Here we consider a binary classification task where only images for digits ``7'' and ``8'' are to be classified. Figure~\ref{vis} shows two amalgamated input images (left two) and their corresponding feature maps obtained at the final hidden layer of the learned model. Note that the top images in the two inputs are identical, but since they are paired with different images to form the input, different features are extracted (e.g., the regions enclosed by the red boxes). This confirms that AgrLearn extracts {\em joint} features across the aggregated examples, as we expect in our design rationale for AgrLearn. 

\subsubsection{Sensitivity to Model Complexity}
With fold-$n$ AgrLearn, the output label space becomes ${\cal Y}^n$. This significantly larger label space seems to suggest that AgrLearn favors more complex model. In this study, we start with AgrLearn-ResNet-18 and investigate the behavior of the model when it becomes more complex. The options we investigate include increasing the model width (by doubling or tripling the number of filters per layer) and increasing the model depth (from 18 layers to 34 layers). The performances of these models are given in   Table~\ref{tab:cap:wide}.

Table~\ref{tab:cap:wide} shows that increasing the model width improves the performance of AgrLearn on both Cifar10 and Cifar100. For example, doubling the number of filters reduces the error rate from 4.73\% to 4.2\% on Cifar10, and tripling the filters further decreases the error rate to 4.14\%. 

\begin{table}[h]
  \centering
\begin{tabular}{l|c|c}\hline
 Data set & Cifar10&Cifar100\\ \hline
 18 layers & 4.73&23.70\\
 18 layers+double&4.20&21.18\\
 18 layers+triple& 4.14&20.21\\
 34 layers&5.01& 21.18\\
 34 layers+double &4.18&22.02\\
\hline
\end{tabular}
  \caption{Test error rates (\%) of AgrLearn-ResNet-18 (``18 layer'') and its more complex variants}
  \label{tab:cap:wide}
\end{table}

However increasing the model depth does not improve its performance as effectively. For example, on  Cifar10, AgrLearn-ResNet-18 (with error rate 4.73\%) outperforms its 34-layer variant (error rate of 5.01\%). But the 34-layer model, when further enhanced by the width, has a performance boost (to error rate 4.18\%).  On Cifar100, increasing 
AgrLearn-ResNet-18 to 34 layers only slightly improves its performance.

We hypothesize that with AgrLearn, the width of a model plays a critical role. This is because the input dimension in AgrLearn increases significantly and the model is 
required to extract joint features across individual objects in the amalgamated example. Nonetheless optimizing over width and optimizing over depth are likely coupled, and they may be further complicated by the internal computations in the model, such as convolution, activation, and pooling.

\subsubsection{Behavior with Respect to Fold Number}
We also conduct experiments investigating the performance of
AgrLearn-ResNet-18 with varying fold $n$ and with respect to varying training sample size $N$.  The Cifar10 dataset is used in this study, as well as two of its randomly reduced subsets, one containing 20\% of the training data and the other containing 50\%. 

\begin{figure}[h]
\vspace{-0.4cm}
    \centering
        \includegraphics[height=2.01528in]{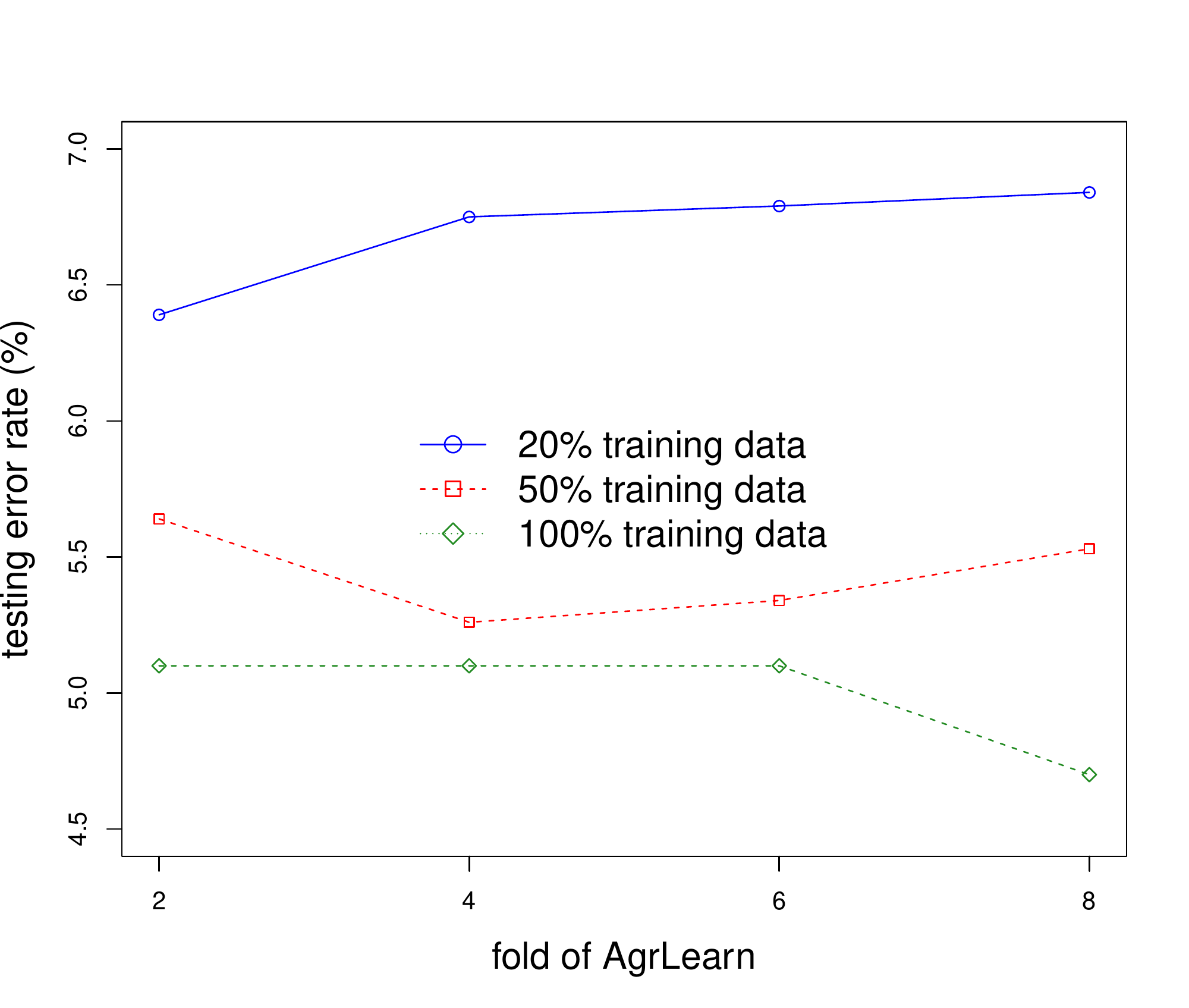}
        \caption{Varying sample sizes and folds}
        \label{fig:performanceTurn}
    \end{figure}

Figure \ref{fig:performanceTurn} suggests that the performance of AgrLearn models vary with 
fold $n$. The best-performing fold number for AgrLearn-ResNet-18 on 20\%, 50\% and 100\%
of the Cifar10 dataset appears to be 2, 4, and 8 (or larger) respectively. This supports our conjecture in Section~\ref{theory} regarding the existence of the critical fold $n^*$.  

\subsubsection{Robustness to Data Scarcity}
\begin{figure}[h]    
       \centering
                \includegraphics[height=2.0158in]{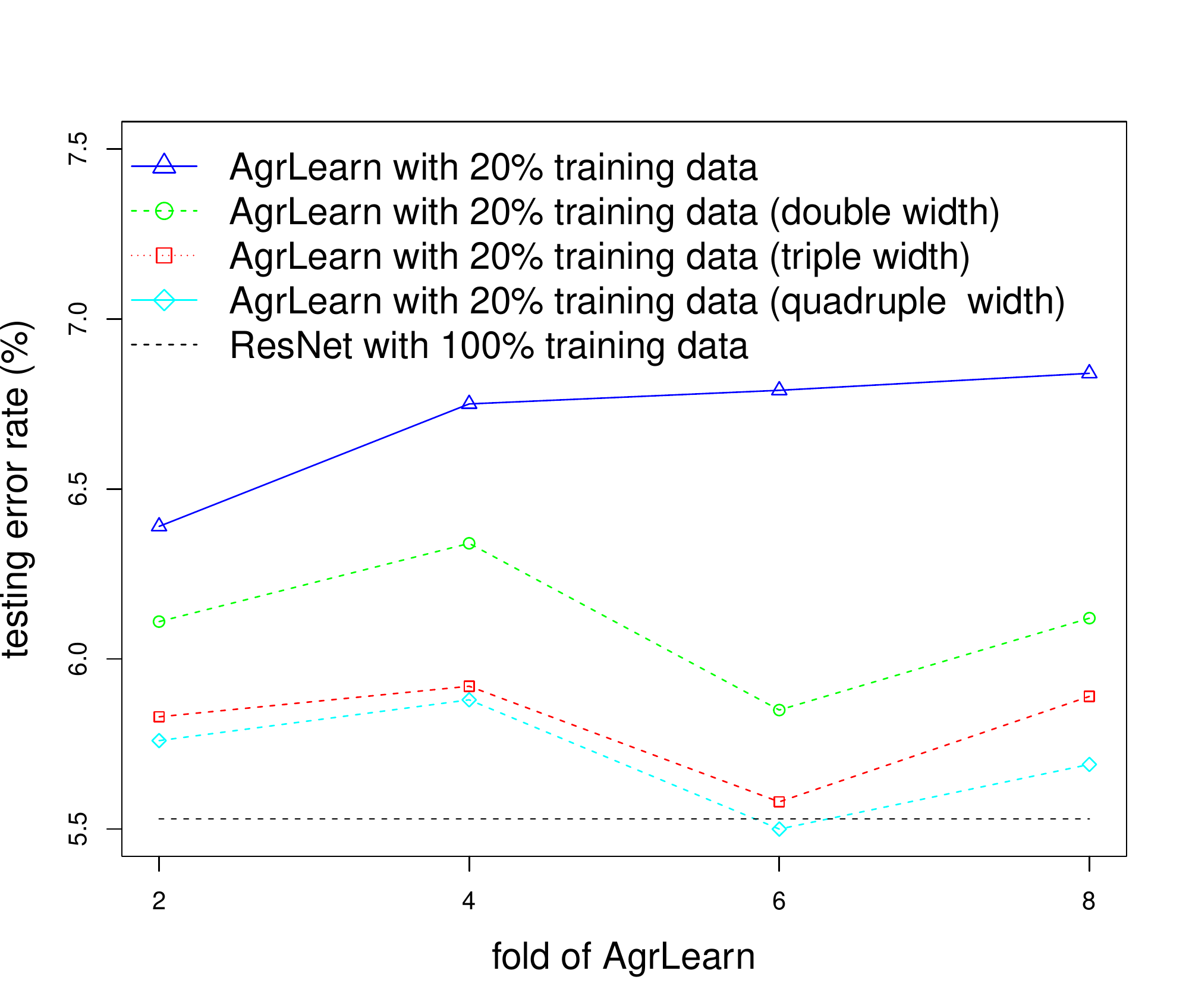}
        \caption{Varying widths and folds, 20\% training examples of Cifar10}
                \label{fig:performanceRobust}
\end{figure}

Using only 20\% of Cifar10, we investigate the performance of AgrLearn-ResNet-18 with increasing widths, namely, with the number of filters doubled, tripled, or quadrupled. In Figure \ref{fig:performanceRobust} we see that the fold-6 triple-width and quadruple-width AgrLearn-ResNet-18 models trained using 20\% of Cifar10 perform comparably to or even better than ResNet-18 trained on the entire Cifar10. This demonstrates the robustness of AgrLearn to data scarcity, making AgrLearn an appealing solution to practical learning problems with inadequate labeled data.

\subsubsection{Impact of Smooth Approximation}
\label{smoother}
Following the discussion in Section~\ref{theory}, we apply a recent data augmentation method  MixUp~\cite{mixup17} as a heuristics to smooth $\left(\widetilde{p}_{XY}\right)^{\otimes n}$.  Briefly, MixUp augments the training data with synthetic training examples, each obtained by interpolating a pair of original examples and their corresponding labels. 
In our experiments, we applied AgrLearn to the same WideResNet architecture as implemented in~\cite{Zagoruyko/code} but only using 22 layers.
Specifically, MixUp is deployed to the 8-fold amalgamated images and their label vectors to augment the training set. 
As shown in Table \ref{tab:sota}, when augmented with smoother MixUp, WideResNet-22-10 model gives rise to a lower error rate of 2.4\% and 16.0\% on Cifar10 and Cifar100, respectively, outperforming the other three state-of-the-art deep models. 

 
 \begin{table}[h]
  \centering
\begin{tabular}{l|c|c}\hline
Dataset & Cifar10& Cifar100  \\\hline 

WideResNet-40-10 &3.8\%&18.3\%\\
DenseNet-BC-190 & 3.7\% & 19.0\% \\
WideResNet-28-10 + MixUp& 2.7\%&17.5\%\\
\hline 
AgrLearn + &&\\
\hspace{0.4cm} WideResNet-22-10 + MixUp &2.4\%&16.0\% \\ 
\hline
\end{tabular}
  \caption{Error rates obtained by various state-of-the-art models; 
Results for  the  WideResNet-40-10 model are from~\cite{DBLP:journals/corr/ZagoruykoK16}; results for the WideResNet-28-10 + MixUp and DenseNet-BC-190 models are  from~\cite{mixup17}; 
  }
  \label{tab:sota}
 \end{table}

\subsection{Text Classification}
\vspace{-0.2cm}
 We test AgrLearn with two widely adopted NLP deep-learning architectures, CNN and LSTM~\cite{Hochreiter:1997:LSM:1246443.1246450}, 
 using two benchmark sentence-classification datasets, Movie Review~\cite{Pang2005}~\footnote{https://www.cs.cornell.edu/people/pabo/movie-review-data/} and Subjectivity~\cite{DBLP:conf/acl/PangL04}. 
 Movie Review and Subjectivity contain respectively 10,662 and 10,000 sentences, with binary labels.  We use 10\% of random examples in each dataset for testing and the rest for training, as is in~\cite{DBLP:conf/emnlp/Kim14}. 
 
For CNN, we adopt CNN-sentence~\cite{DBLP:conf/emnlp/Kim14} and implement it exactly as in~\cite{kim/code}. For LSTM, we 
just simply replace the convolution and pooling components in CNN-sentence with standard LSTM units as implemented 
in~\cite{Abadi:2016:TSL:3026877.3026899}. The final feature map of CNN and final state of LSTM are passed to a logistic regression classifier for label prediction. 
Each sentence enters the models via a learnable, 
randomly initialized word-embedding dictionary. For CNN, all sentences are zero-padded to the same length.

The fold-2 AgrLearn models corresponding to the CNN and LSTM models are also constructed. In AgrLearn-CNN, the aggregation of two sentences in each input simply involves concatenating 
the two zero-padded sentences. In AgrLearn-LSTM, when two sentences are concatenated in tandem, an EOS word is inserted after the first sentence. 

We train and test the CNN, LSTM and their respective AgrLearn models on the two datasets, and report their performances in Table~\ref{tab:accuracy:cnn}. Clearly, the AgrLearn models improve upon their corresponding CNN or LSTM counterparts. In particular, the relative performance gain brought by AgrLearn on the CNN model appears more significant, amounting to 4.2\% on Movie Review and 3.8\% on Subjectivity.

\begin{table}[H]
  \centering
\begin{tabular}{l|c|c}\hline
Dataset & CNN& AgrLearn-CNN   \\\hline 
Movie Review&76.1 &79.3\\
Subjectivity&90.01&93.5\\ \hline \hline
Dataset& LSTM& AgrLearn-LSTM  \\\hline 
Movie Review&76.2 &77.8\\
Subjectivity&90.2&92.1 \\

\hline
\end{tabular}
  \caption{Accuracy (\%) obtained by CNN, LSTM and their respective AgrLearn models.}
  \label{tab:accuracy:cnn}
  \vspace{-0.5cm}
\end{table}

\section{Conclusion, Discussion and Outlook}
\label{future}

Aggregated Learning, or AgrLearn, is a simple and effective neural network modeling framework, justified information-theoretically. It builds on an equivalence  between IB learning and IB quantization and exploits the power of vector quantization, well known in information theory. As shown in our experiments, AgrLearn can be applied to an existing network architecture with virtually no programming or tuning effort. We have demonstrated its effectiveness through the significant performance gain it brings to the current art of deep network models. 
Its robustness to small training samples is also a salient feature, making AgrLearn particularly  attractive in practice, where the labeled data may not be abundant. 

We need to acknowledge that the gain brought by AgrLearn is not free of cost. In fact, the memory consumption of fold-$n$ AgrLearn is $n$ times what is required by the conventional model. 


Another line of research, seemingly related to AgrLearn, is data augmentation  ~\cite{Simard98transformationinvariance,DBLP:journals/corr/abs-1806-05236,mixup17}, including, for example, MixUp. In our opinion, however, it is incorrect to regard AgrLearn as data augmentation. Aggregating examples in AgrLearn expands the input space, which induces additional modeling freedom (although we made little effort in this work exploiting this freedom). Such a  property, not possessed by data augmentation schemes, clearly distinguishes AgrLearn from those schemes. 

The motivation of AgrLearn is the fundamental notion of information bottleneck (IB). The effectiveness of AgrLearn demonstrated in this paper may serve as additional validation of the IB theory. 

We believe that the proposal and successful application of AgrLearn in this paper are believed to signal the beginning of a promising and rich theme of research. 
Many interesting questions deserve further investigations. For example,  how can we characterize the interaction between model complexity, fold number and sample size in AgrLearn? how can the modeling freedom in AgrLearn be fully exploited? And finally, how can the true IB-learning objective (\ref{eq:IB-learn-problem}) be more accurately and yet efficiently approximated?

\clearpage



\bibliography{reference}
\bibliographystyle{icml2019}

\clearpage

\onecolumn



\section*{Supplementary material}

# Supplementary Materials

## A Preliminaries

Here we give a brief review of typical sequences [5], which will be useful in proving Theorem 1. We remark that the notion of typicality here is stronger than the widely used (weak) typicality in [2]. and refer the interested reader to [3] for a comprehensive treatment of the subject. Throughout this note, the symbol "$E$" will denote expectation. At some places, we might use subscripts to explicitly indicate the random variables with respect to which the expectation is performed.

1. <u>Empirical distribution</u>: Given a sequence $x^n \in \mathcal{X}^n$, it induces an empirical distribution on $\mathcal{X}$ defined as

$$\pi(x|x^n) := \frac{1}{n}|\{i : x_i = x\}| \text{ for all } x \in \mathcal{X}. \tag{A.1}$$

2. <u>Typical set</u>: For $X \sim p_X(x)$ and $\epsilon \in (0, 1)$, the set of $\epsilon$-typical sequences is defined as

$$\mathcal{S}_\epsilon^n(X) := \{x^n \mid |\pi(x|x^n) - p_X(x)| \leq \epsilon p_X(x) \text{ for all } x \in \mathcal{X}\}. \tag{A.2}$$

3. <u>Typical average lemma</u>: For any $x^n \in \mathcal{S}_\epsilon^n(X)$ and any non-negative function $g$ on $\mathcal{X}$, we have

$$(1 - \epsilon)E[g(X)] \leq \frac{1}{n}\sum_i g(x_i) \leq (1 + \epsilon)E[g(X)]. \tag{A.3}$$

Note that by choosing $g$ to be the $\log$ function, one recovers the notion of typicality in [2]. The typicality here is strictly stronger than the one in [2], however, similar to weak typicality, most i.i.d. sequences are still typical under this definition. Namely, for any i.i.d sequence $X^n$ of RVs with $X_i \sim p_X(x_i)$, by the LLN, the empirical distribution $\pi(x|X^n)$ converges (in probability) to $p_X(x)$, for all $x \in \mathcal{X}$, and so such sequence, with high probability, belongs to the typical set.

4. <u>Joint typicality</u>: Items 1 and 2 extend to a joint source $(X, Y) \sim p_{XY}(x, y)$ in the obvious way, i.e., by treating $X$ and $Y$ as one source $(X, Y)$. Given a sequence $(x^n, y^n) \in \mathcal{X}^n \times \mathcal{Y}^n$, it induces an empirical distribution on $\mathcal{X} \times \mathcal{Y}$ defined as

$$\pi(x, y|x^n, y^n) := \frac{1}{n}|\{i : x_i = x, y_i = y\}| \text{ for all } (x, y) \in \mathcal{X} \times \mathcal{Y}. \tag{A.4}$$

For $X \sim p_X(x)$ and $\epsilon \in (0, 1)$, the set of $\epsilon$-typical sequences is defined as

$$\mathcal{S}_\epsilon^n(X, Y) := \{(x^n, y^n) \mid |\pi(x, y|x^n, y^n) - p_{XY}(x, y)| \leq \epsilon p_{XY}(x, y) \text{ for all } (x, y) \in \mathcal{X} \times \mathcal{Y}\}. \tag{A.5}$$

5. <u>Joint typicality lemma</u>: Let $(X, Y) \sim p_{XY}(x, y)$ and $p_Y(y)$ be the marginal distribution $\sum_x p_{XY}(x, y)$. Then, for $\epsilon' < \epsilon$, there exists $\delta(\epsilon) \to 0$ as $\epsilon \to 0$ such that

$$p\{(x^n, Y^n) \in \mathcal{S}_\epsilon^n(X, Y)\} \geq 2^{-n(I(X;Y)+\delta(\epsilon))}. \tag{A.6}$$

for $x^n \in \mathcal{S}_{\epsilon'}^n$, $Y^n \sim \prod_{i=1}^n p_Y(y_i)$, and sufficiently large $n$,

# B  Proof of Theorem 1

Before we present a proof, we pause and make few remarks. The proof follows standard techniques from information theory for proving results of this nature. It is worth noting that the conventional proof of achievability [2] of the rate-distortion theorem does not directly apply here since the distortion measure $d_{\text{IB}}$ depends on the distribution $p_{T|X}$. This was addressed in [4] by extending the definition of distortion jointly typical sequences in [2] to multi-distortion jointly typical sequences. Our approach exploits the notion of typicallity presented in the previous section and closely follows the proof of achievability in [3] of the rate-distortion theorem.

$$R'(D) := \min_{p_{T|X}(t|x): E[d(X,T)] \leq D} I(X;T). \tag{B.1}$$

We need to show $R_{\text{IB-RD}}(D) = R'(D)$.

## B.1  Proof of the converse

We first show $R_{\text{IB-RD}}(D) \geq R'(D)$ by showing that for any sequence of $(2^{nR}, n)$ codes satisfying (7), it must be the case that $R \geq R'(D)$. We have

$$nR \overset{(i)}{\geq} H(f_n(X^n)) \overset{(ii)}{\geq} I(X^n; f_n(X^n)) \overset{(iii)}{\geq} I(X^n, T^n)$$
$$= \sum_i H(X_i) - H(T_i | X^n, T^{i-1})$$
$$\geq \sum_i H(X_i) - H(T_i | X_i)$$
$$= \sum_i I(X_i; T_i)$$
$$\overset{(iv)}{\geq} \sum_i R'(E[d(X_i, T_i)])$$
$$\overset{(v)}{\geq} nR'\left(\frac{1}{n} \sum_i E[d(X_i, T_i)]\right)$$
$$\overset{(vi)}{=} nR'(E[d(X^n, T^n)])$$
$$\overset{(vii)}{\geq} nR'(D),$$

where (i) follows from the fact that $f_n$ takes its values from $\{1, \ldots, 2^n\}$, (ii) from the non-negativity of conditional entropy, (iii) from the data processing inequality since $T^n = g_n(f_n(X^n))$, (iv) from (B.1) by noting that $R'(E[d(X_i, T_i)]) = \min_{p_{T_i|X_i}} I(X_i; T_i)$, (vi) from (6), and (vii) from (7) since $R'(D)$ is a decreasing function in $D$. To prove (v), it is sufficient to show that $R'$ is a convex function in $D$, which is shown in the following lemma. $\square$

**Lemma 1** ([1, Lemma 1]). *The function $R'(D)$ defined in* (B.1) *is a convex function.*

*Proof.* Let $(D_1, R_1)$ and $(D_2, R_2)$ be two points on $R'(D)$ attained, respectively, by $T_1$ and $T_2$ via the minimizers $p_{T_1|X}$ and $p_{T_2|X}$ of (B.1). Define

$$T = \begin{cases} T_1, & Z = 1 \\ T_2, & Z = 2, \end{cases}$$

where $Z \in \{1, 2\}$ is a RV independent of $(T_1, T_2, X, Y)$ with $p_Z(1) = \lambda$. Then,

$$p_{XTZ}(x, t, z) = \begin{cases} \lambda \cdot p_{XT_1}(x, t), & Z = 1 \\ (1-\lambda) \cdot p_{XT_2}(x, t), & Z = 2, \end{cases}$$

and so

$$I(X;T,Z) = \sum_{x,t,z} p_{XTZ}(x,t,z) \log \frac{p_{XTZ}(x,t,z)}{p_X(x)p_{TZ}(t,z)}$$

$$= \sum_{x,t} \lambda \cdot p_{XT_1}(x,t) \log \frac{\lambda \cdot p_{XT_1}(x,t)}{\lambda \cdot p_X(x)p_{T_1}(t)} + \sum_{x,t} (1-\lambda) \cdot p_{XT_2}(x,t) \log \frac{(1-\lambda) \cdot p_{XT_2}(x,t)}{(1-\lambda) \cdot p_X(x)p_{T_2}(t)}$$

$$= \lambda \cdot I(X;T_1) + (1-\lambda) \cdot I(X;T_2).$$

Moreover, we have

$$E[d(X,(T,Z))] = \sum_{x,t,z} p_{XTZ}(x,t,z) \sum_y p_{Y|X}(y|x) \log \frac{q_{Y|X}(y|X)}{q_{Y|TZ}(y|t,z)}$$

$$= H(Y|TZ) - H(Y|X)$$

$$= \lambda \cdot H(Y|T_1) + (1-\lambda) \cdot H(Y|T_2) - \lambda \cdot H(Y|X) - (1-\lambda) \cdot H(Y|X)$$

$$= \lambda \cdot E[d(X,T_1)] + (1-\lambda) \cdot E[d(X,T_2)].$$

Since $(T,Z)$—$X$—$Y$ is a markov chain resulting in cost and constraint that are linear functions of the original costs and constraints, the claim follows from the definition of $R'$ in (B.1). $\square$

## B.2 Proof of Achievability in Theorem 1

We need to show that for $R = R'(D)$ there exists a sequence $(2^{nR}, n)$ of codes satisfying (7).

<u>Random codebook:</u> Let $R = R'(D)$ and fix $p_{T|X}$ to be an optimal distribution to the minimization (B.1) at $D/(1+\epsilon)$, i.e., we pick a conditional distribution that attains $R'(D/(1+\epsilon))$.[1] Let $p_T(t) = \sum_{x \in \mathcal{X}} p_X(x)p_{T|X}(t|x)$. Generate $2^{nR}$ i.i.d. sequences $t^n(m) \sim \prod_{i=1}^n p_T(t_i)$, $m \in \{1, \ldots, 2^{nR}\}$. These sequences form the codebook which is revealed to the encoder and decoder.

<u>Encoder:</u> The encoder uses joint typicality encoding. Given a sequence $x^n$, find an index $m$ s.t. $(x^n, t^n(m)) \in \mathcal{S}_\epsilon^n(X,T)$ and send $m$. If there is more than one index then choose $m$ to be the smallest index, and if there is no index then choose $m = 1$. (In other words, the encoder sets $f_n(x^n)$ to be the index $m$, where $m$ is as described above.)

<u>Decoder:</u> Upon receiving index $m$, set $t^n = t^n(m)$. (In other words, the decoder sets $g_n(m)$ to be the row of the codebook indexed by $m$.)

<u>Expected distortion</u> Let $\epsilon' < \epsilon$ and $M$ be the index chosen by the encoder. We first bound the distortion averaged over codebooks. Towards this end, define the event

$$\mathcal{E} := \{(X^n, T^n(m)) \notin \mathcal{S}_\epsilon^n(X,T)\},$$

then by the union bound and the choice of the encoder, we have

$$p(\mathcal{E}) \leq p(\mathcal{E}_1) + p(\mathcal{E}_2),$$

where

$$\mathcal{E}_1 := \{X^n \notin \mathcal{S}_{\epsilon'}^n(X)\},$$
$$\mathcal{E}_2 := \{X^n \in \mathcal{S}_{\epsilon'}^n, (X^n, T^n(m)) \notin \mathcal{S}_\epsilon^n(X,T) \,\forall m \in \{1, \ldots, 2^{nR}\}\}.$$

---

[1] A comment on existence. There is a feasible distribution $p_{T|X}$ satisfying the distortion constraint for any $D$. For $D = 0$, choose $p_{T|X}(t|x) = p_X(t)$ and for $D \geq D_{\max} := I(X;Y)$ choose $p_{T|X}$ as the degenerate distribution that assigns all the weight on one element of $T$. For $D \in [0, D_{\max}]$, use a latent variable $Z$ as in the proof of the converse with $\lambda = D/D_{\max}$.

3

We have $\lim_{n\to\infty} p(\mathcal{E}_1) = 0$ by the LLN and

$$p(\mathcal{E}_2) = \sum_{x^n \in \mathcal{S}_{\epsilon'}^n} p_{X^n}(x^n) p\{(x^n, T^n(m)) \notin \mathcal{S}_\epsilon^n \,\forall m \mid X^n = x^n)\}$$

$$\stackrel{(i)}{=} \sum_{x^n \in \mathcal{S}_{\epsilon'}^n} p_{X^n}(x^n) \prod_{m=1}^{2^{nR}} p\{(x^n, T^n(m)) \notin \mathcal{S}_\epsilon^n\}$$

$$\stackrel{(ii)}{=} \sum_{x^n \in \mathcal{S}_{\epsilon'}^n} p_{X^n}(x^n) \big(p\{(x^n, T^n(1)) \notin \mathcal{S}_\epsilon^n\}\big)^{2^{nR}}$$

$$\stackrel{(iii)}{\leq} \sum_{x^n \in \mathcal{S}_{\epsilon'}^n} p_{X^n}(x^n) \big(1 - 2^{-nI(X;T)+\delta(\epsilon)}\big)^{2^{nR}}$$

$$\leq \big(1 - 2^{-nI(X;T)+\delta(\epsilon)}\big)^{2^{nR}}$$

$$\stackrel{(iv)}{\leq} \exp\big(-2^{n(R-I(X;T)-\delta(\epsilon))}\big),$$

where (i) and (ii) are by the i.i.d assumption on the codewords, (iii) is by the joint typicality lemma (A.6), (iv) is by the fact $(1-\alpha)^k \leq \exp(-k\alpha)$ for $\alpha \in [0,1]$ and $k \geq 0$. Hence, we have $\lim_{n\to\infty} p(\mathcal{E}_2) = 0$ for $R > I(X;T) + \delta(\epsilon)$.

Now, the distortion averaged over $X^n$ and over the random choice of the codebook is given as

$$E_{X^n,T^n,M}[d(X^n, T^n(M))] = p(\mathcal{E}) \cdot E_{X^n,T^n,M}[d(X^n, T^n(M))|\mathcal{E}] + p(\mathcal{E}^c) \cdot E_{X^n,T^n,M}[d(X^n, T^n(M))|\mathcal{E}^c]$$
$$\leq p(\mathcal{E}) \cdot d_{\max} + p(\mathcal{E}^c) \cdot E_{X^n,T^n,M}[d(X^n, T^n(M))|\mathcal{E}^c]$$
$$= p(\mathcal{E}) \cdot d_{\max} + p(\mathcal{E}^c) \cdot E_{X^n,T^n}[d(X^n, T^n(1))|\mathcal{E}^c]$$
$$\leq p(\mathcal{E}) \cdot d_{\max} + p(\mathcal{E}^c) \cdot (1+\epsilon) \cdot E_{X,T}[d(X,T)],$$

where $d_{\max} = \max_{(x,t) \in \mathcal{X} \times \mathcal{T}} d(x,t)$. By the choice of $p_{T|X}(t|x)$, we have $E[d(X,T)] \leq D/(1+\epsilon)$, and so

$$\lim_{n\to\infty} E_{X^n,T^n,M}[d(X^n, T^n(M))] \leq D,$$

for $R > I(X,T) + \delta(\epsilon)$, where $\delta(\epsilon) \to 0$ as $n \to \infty$. Since the expected distortion, averaged over codebooks, satisfies the distortion constraint $D$, there must exist a sequence of codes that satisfies the constraint. This shows the achievability of the rate-distortion pair $(R(D/(1+\epsilon)) + \delta(\epsilon), D)$. By the continuety of $R(D)$ in $D$ the achievable rate $R(D/(1+\epsilon)) + \delta(\epsilon)$ converges to $R(D)$ as $\epsilon \to 0$. $\square$

## C  Proof of Theorem 2

Follows directly from Theorem 1 and the definition of $d_{\text{IB}}$. Namely, we have

$$E_{X,\hat{X}}(d_{\text{IB}}(X,\hat{X})) := \sum_{x,\hat{x}} d_{\text{IB}}(x,\hat{x}) p_{X\hat{X}}(x,\hat{x})$$
$$= I(X,Y) - I(\hat{X},Y), \tag{C.1}$$

where the second equality is by the definition of $d_{\text{IB}}$ and the Markov chain $T$—$X$—$Y$. Hence, we have

$$R_{\text{IB}-\text{RD}}(D) = \min_{p_{T|X}(t|x): E[d(X,T)] \leq D} I(X;T)$$
$$= \min_{p_{T|X}(t|x): I(Y;T) \geq I(X;Y)-D} I(X;T)$$
$$= R_{\text{IB}-\text{learn}}(I(X;Y) - D),$$

where the first equality is by Theorem 1, the second by (C.1), and the third by the definition of $R_{\text{IB}-\text{learn}}$ in (2). The claim follows via the substitution $A := I(X;Y) - D$. $\square$

4



\end{document}